\documentclass[letterpaper, 10 pt, conference]{ieeeconf}  %

\IEEEoverridecommandlockouts                              %

\usepackage{graphics} %
\usepackage{epsfig} %
\usepackage{mathptmx} %
\usepackage{times} %
\usepackage{amsmath} %
\usepackage{amssymb}  %
\usepackage[utf8]{inputenc} %
\usepackage[T1]{fontenc}    %
\usepackage{hyperref}       %
\usepackage{url}            %
\usepackage{booktabs}       %
\usepackage{amsfonts}       %
\usepackage{nicefrac}       %
\usepackage{microtype}      %
\usepackage{xcolor}         %
\usepackage{graphicx}
\usepackage{xspace}
\usepackage{mathtools}  %
\usepackage{pifont}%
\usepackage{multirow}
\usepackage{wrapfig}
\usepackage{array}
\usepackage{tabularx}
\usepackage{soul}

\newcommand{\mypar}[1]{\vspace{1mm}\noindent\textbf{#1}}

\def\eg{\textit{e.g.}\@\xspace}
\def\ie{\textit{i.e.}\@\xspace}

\DeclareMathOperator*{\argmin}{arg\,min}
\newcommand\norm[1]{\lVert#1\rVert}
\newcommand{\cmark}{\ding{51}}%
\newcommand{\xmark}{\ding{55}}%
\newcommand{\best}[1]{\textbf{#1}}
\newcommand{\secondbest}[1]{\underline{#1}}
\newcommand{\bC}{\mathbf{C}}
\newcommand{\hbx}{\hat{\mathbf{x}}}
\newcommand{\hby}{\hat{\mathbf{y}}}

\title{\LARGE \bf
Direct Superpoints Matching for Robust Point Cloud Registration
}
\title{\LARGE \bf A Strong Baseline for Point Cloud Registration \\via Direct Superpoints Matching}

\author{Aniket Gupta$^{1}$, Yiming Xie$^{1}$, Huaizu Jiang$^{1}$, Hanumant Singh$^{2}$%
\thanks{$^{1}$Khoury College of Computer Science, Northeastern University}%
\thanks{$^{2}$College of Engineering, Northeastern University
        {\tt\small \{gupta.anik, xie.yim, ha.singh, h.jiang\}@northeastern.edu}}%
}

\begin{document}

\maketitle
\thispagestyle{empty}
\pagestyle{empty}

\begin{abstract}

Deep neural networks endow the downsampled superpoints with highly discriminative feature representations. Previous dominant point cloud registration approaches match these feature representations as the first step, \eg, using the Sinkhorn algorithm. A RANSAC-like method is then usually adopted as a post-processing refinement to filter the outliers. Other dominant method is to directly predict the superpoint matchings using learned MLP layers. Both of them have drawbacks: RANSAC-based methods are computationally intensive and prediction-based methods suffer from outputing non-existing points in the point cloud.
In this paper, 
we propose a straightforward and effective baseline to find correspondences of superpoints in a global matching manner.
We employ the normalized matching scores as weights for each correspondence, allowing us to reject the outliers and further weigh the rest inliers when fitting the transformation matrix without relying on the cumbersome RANSAC. 
Moreover, the entire model can be trained in an end-to-end fashion, leading to better accuracy. 
Our simple yet effective baseline shows comparable or even better results than  state-of-the-art methods on three datasets including ModelNet, 3DMatch, and KITTI.
We do not advocate our approach to be \emph{the} solution for point cloud registration but use the results to emphasize the role of matching strategy for point cloud registration.
The code and models are available at \href{https://github.com/neu-vi/Superpoints_Registration}{https://github.com/neu-vi/Superpoints\_Registration}

\end{abstract}

\section{Introduction}

Point cloud registration refers to the task of aligning two partially overlapping point clouds into a shared coordinate system. 
In this paper, we tackle this problem where the goal is to determine the transformation matrix, including rotation and translation, from one point cloud (source) to the other (target). 
It has attracted a lot of research interest due to its broad applications in SLAM (Simultaneous Localization and Mapping)~\cite{Deschaud2018imlsslam,legoloam2018}, autonomous driving~\cite{li2021deep,Lu2019L3NetTL}, 3D reconstruction~\cite{Izadi2011kinect,Gross193DV}, etc.

\begin{figure}[t]
    \centering
       \includegraphics[width=1\linewidth]{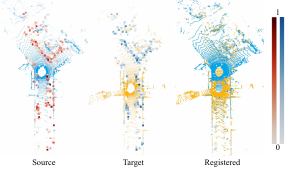}
       \caption{
           \textbf{Our approach directly matches the superpoints between the input point clouds to register them by estimating the $SE(3)$ transformation matrix}. The correlation weights obtained from matching are used to filter out incorrect correspondences (outliers) and further weigh the rest inliers for the transformation matrix estimation.
           The source cloud is shown in blue, with superpoints highlighted in red. Similarly, the target point cloud is displayed in yellow, with matching superpoints shown in blue. 
           The intensity of the red and blue colors represents the correspondence weights. Notice how most keypoints are distributed around the three-way junction and exhibit high correspondence weights. (Best viewed in color.)
           }
       \label{fig:teaser}
\end{figure}

A prevailing paradigm to solve the registration task is to leverage the correspondences of \emph{superpoints} across two point clouds, which can be obtained using either keypoint detectors~\cite{ao2021spinnet,bai2020d3feat,choy2019fully,huang2021predator} or downsampling in deep neural networks~\cite{Qi2017piontnet,Qi2017pointnetpp,thomas2019kpconv}.
Ideally, these superpoints should capture salient and distinctive points or regions within a point cloud. 
With the learned feature representations, the superpoints are endowed with sufficient discriminative power so they can be matched across the source and target point clouds, where the matching strategy is crucial for finding correspondences of superpoints and can significantly affect the convergence and performance of the network. 
Previous approaches ~\cite{yew2020rpm, qin2022geometric} use the Sinkhorn algorithm~\cite{sinkhorn1967concerning} to match superpoints, which however is sensitive to initialization parameters and requires careful tuning. 
In ~\cite{sun2021loftr}, Dual-Softmax is adopted for feature matching between two input images by using kepoints which have softmax correlation scores greater than a certain threshold. 
The correspondences of superpoints, however, are inevitably noisy due to either the noisy sensory data or simply incorrect estimations of correspondences, which prevent existing approaches from directly matching superpoints for point cloud registration.
Postprocessing refinement is usually needed, for instance, using RANSAC-like approaches to prune the outliers~\cite{choy2019fully, deng2018ppf, zeng20173dmatch,yu2021cofinet}. 
But such refinement is inherently slow due to the iterative nature of the RANSAC pipeline and can not be easily integrated into an end-to-end trainable system.

The recent work~\cite{yew2022regtr} populates another strategy, which uses a two-layer MLP to direct predict matched superpoints between the input point clouds.
But such predicted correspondences may not exist in the point cloud, leading to inferior registration accuracy.

In this paper, we present a \emph{strong baseline} for point cloud registration by directly matching superpoints.
Specifically, building upon RegTR \cite{yew2022regtr}, instead of using a MLP to predict correspondences of superpoints, we find the correspondences by computing the similarity scores of all superpoints across the source and target point cloud in a global matching manner.
Their normlized matching scores can be used to 
filter out the unreliable correspondences (\ie, outliers). 
By integrating the weights of the rest inlier superpoint matchings into a differentiable variant of the Kabsh-Umeyama Algorithm \cite{88573, kabsch1976solution}, we obtain robust estimations of the $SE(3)$ transformation between the input point cloud pairs.
An illustration is shown in Fig.~\ref{fig:teaser}.
As a result, no ad hoc postprocessing refinement is needed, yielding a more efficient model.
More importantly, the entire model can be trained in an end-to-end manner, where the feature representation learning, superpoints matching, and transformation estimation can be jointly optimized. 
Better registration accuracy can thus be obtained.
In addition, compared to~\cite{yew2022regtr}, our approach find correspondences of superpoints by global matching, which does not output non-existing points in the target point cloud.

We run experiments on three benchmark datasets, including ModelNet~\cite{wu20153d}, 3DMatch~\cite{zeng20173dmatch}, and KITTI \cite{Geiger2012CVPR}, and achieve comparable or even better results than state-of-the-art methods. 
We do not advocate our approach to be \emph{the} solution for point cloud registration.
Rather, we'd like to emphasize the role of matching strategy for point cloud registration, showcasing the feasibility of achieving high accuracy without cumbersome ad hoc postprocessing.
By releasing the code and model weights, we hope our work can foster future research.

\begin{figure*}[t]
    \centering
       \includegraphics[width=1\linewidth]{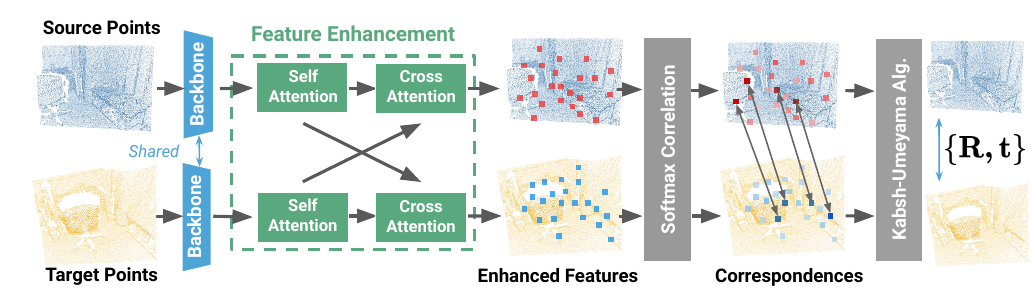}
       \caption{
           \textbf{Model Architecture}: The KPConv backbone downsamples the input point cloud and generates superpoints and their feature vectors. These superpoint features are then conditioned on the other point cloud in the feature enhancement block. Lastly, superpoint features can be directly matched using Global Softmax to estimate SE(3) transformation using the corrleation weights in a robust manner. The intensity
of the red and blue colors of the superpoints after the softmax correlation step represents the correspondence weights. (Best viewed in color.)
           }
       \label{fig:arch}
       \vspace{-2mm}
\end{figure*}
\section{Related Work}

\mypar{Traditional registration approaches.}
The most known algorithm Iterative Closest Point (ICP)~\cite{besl1992method} has been widely used for point cloud registration. 
ICP solves the registration problem iteratively in two steps: (1) It obtains the spatially closest point correspondence and then (2) finds the least-squares rigid transformation. 
The spatial-distance-based correspondences are sensitive to the initial transformation and point noises. 
A lot of variants~\cite{rusinkiewicz2001efficient, chetverikov2002trimmed, segal2009generalized, bouaziz2013sparse, rusinkiewicz2019symmetric,liu2018efficient} have been proposed to improve ICP. 
Another method is to extract and match keypoints based on feature extraction methods such as FPFH~\cite{rusu2009fast} and SHOT~\cite{tombari2010unique}, followed by an outlier rejection postprocessing step.

\mypar{Learning-based registration approaches.}
Recently, many works have used deep learning for point cloud learning and registration. 
Some work first estimates the correspondence between two point clouds and then computes the transformation with some robust pose estimators. 
To predict the correspondence between two point clouds, 3DMatch~\cite{zeng20173dmatch} detects the repeatable keypoints and learns discriminative descriptors for keypoints. 
The following works aim to either improve the keypoint detections~\cite{bai2020d3feat,li2019usip,yew20183dfeat}
or learn better feature descriptors~\cite{choy2019fully, deng2018ppf, deng2018ppfnet, KhouryLCGF, ao2021spinnet, wang2022you}. 
Predator~\cite{huang2021predator} uses the attention mechanism proposed in Transformers~\cite{vaswani2017attention} to enhance the point feature descriptors. 
Other detector-free methods~\cite{yu2021cofinet, qin2022geometric} extract the correspondences by considering all possible matches. 
Another line of work~\cite{cao2021pcam, yuan2020deepgmr} has included the transformation computation into the training pipeline. 
Unlike these works which require either ad-hoc postprocessing or coarse-to-fine registration, our method directly matches the superpoints without any refinement.

\mypar{Learning visual correspondence.} Various approaches have been proposed to establish correspondences between the input, \eg, using nearest neighbor followed by the distance ratio test \cite{SIFT}.
Recent approaches~\cite{yew2020rpm, qin2022geometric} use the Sinkhorn algorithm~\cite{sinkhorn1967concerning} to match superpoints for point cloud registration. However, it is sensitive to initialization parameters and requires careful tuning. 
In ~\cite{sun2021loftr}, Dual-Softmax is adopted for feature matching between two input images by using kepoints which have softmax correlation scores greater than a certain threshold. 
Global Softmax is used in a recent work of optical flow~\cite{xu2022gmflow} to find global matchings by simply taking the correspondence for each pixel with the highest correlation score.
In this paper, we extend Global Softmax by using the corrleation weights for both filtering unreliable correspondences and weighing the remaining inliers when estimating the SE(3) transformation matrix.

\mypar{Correspondence filters.}
RANSAC~\cite{fischler1981random} is typically used to filter out the outliers in the predicted correspondence to obtain a robust transformation estimation. 
However, RANSAC is relatively slow and cannot be incorporated into the training pipeline because the hypothesis selection step is non-differentiable.
To alleviate these problems, DSAC~\cite{brachmann2017dsac} and $\nabla$-RANSAC~\cite{wei2023generalized} modify the RANSAC pipeline and make it differentiable. But even the differentiable versions are similar in computational complexity as vanilla RANSAC. %
Other deep robust estimators~\cite{bai2021pointdsc, choy2020deep, pais20203dregnet, gojcic2020learning, lee2021deep, yi2018learning} usually use the classification network to identify which correspondences are outliers and then reject them.
Instead of using these complex correspondence filters, our method can directly filter out outliers effectively by leveraging the rich information in the superpoints matching.

\section{Method}
Given the source and target point clouds $ \mathbf{X} \in \mathbb{R}^{M\times3} $ and $\mathbf{Y} \in \mathbb{R}^{N\times3}$, our goal is to determine the SE(3) transformation $\mathbf{T} = \{\mathbf{R}, \mathbf{t}\}$ with rotation $\mathbf{R} \in SO(3)$ and translation $\mathbf{t} \in \mathbb{R}^3$ to align two point clouds into a common coordinate system.
$M$ and $N$ denote the numbers of points.

\newcommand{\bX}{\mathbf{X}}
\newcommand{\bY}{\mathbf{Y}}
\newcommand{\bXp}{\mathbf{X}{'}}
\newcommand{\bYp}{\mathbf{Y}{'}}
\newcommand{\bF}{\mathbf{F}}

\subsection{Superpoints Feature Extraction and Enhancement}
Following \cite{yew2022regtr}, we use Kernel Point Convolution (KPConv)~\cite{thomas2019kpconv} as the backbone to selectively downsample the point cloud into a set of superpoints and extract global feature vectors for each superpoint. 
The KPConv backbone uses a series of ResNet-like blocks~\cite{he2016deep} and convolutions to downsample the input point clouds into a reduced set of superpoints $\bXp \in \mathbb{R}^{M'\times3}$ and $\bYp \in \mathbb{R}^{N'\times3}$, where $M' < M$ and $N' < N$.
The superpoints are described by their feature vectors $\bF_{\bXp} \in \mathbb{R}^{M'\times D}$ and $\bF_{\bYp} \in \mathbb{R}^{N'\times D}$, respectively, with $D$ being the feature dimension. 
The network weights are shared among the two point clouds.
We use a shallower backbone for 3DMatch dataset compared to \cite{yew2022regtr, huang2021predator} to avoid significant downsampling by removing the 4-th residual block.

Although KPConv backbone provides reasonably good representations, the superpoints features are obtained within each point cloud \emph{independently}. To obtain highly discriminative feature representations for superpoints matching, We use the multi-head attention mechanism in the Transformer model~\cite{vaswani2017attention} as the feature enhancement module as suggested in \cite{yew2022regtr}, shown in Fig.~\ref{fig:arch}.
It consists of both self and cross-attention, where the self-attention is to integrate the information from the other points within the same point cloud  and the cross-attention allows interactions with  points in another point cloud to consider the mutual dependencies.
In addition to the multi-head attention, other components in the Transformer model, including position encodings of 3D points, residual connections, layer normalization, and feed-forward network are applied to each layer.
The entire feature enhancement module consists of 6 such layers with 256 dimensions and 8 attention heads.

\newcommand{\bbF}{\bar{\mathbf{F}}}
The outputs of the feature enhancement module are features $\bbF_{\bXp} \in \mathbb{R}^{M'\times D}$ and $\bbF_{\bYp} \in \mathbb{R}^{N'\times D}$ which has aggregated geometric information from both source and target point cloud. The strongly associated features are strengthened while the weakly associated features are weakened.

\subsection{Superpoint Matching for SE(3) Transformation Estimation}
Unlike \cite{yew2022regtr} that uses a two-layer MLP to predict corresponding points, we find the correspondences of superpoints in a global matching manner.
This change removes the ambiguity of predicting non-existing corresponding points.
To get the correspondences between two point clouds, we first compare the feature similarity for each point in $\bbF_{\bXp}$ to all points in $\bbF_{\bYp}$ by computing their correlations~\cite{xu2022gmflow}, which can be done efficiently in a single step as follows:
\begin{equation}
    {\mathbf{C}} =  \mathrm{Softmax}\left(\bbF_{\bXp} \cdot \bbF_{\bYp}^T\right)  \in \mathbb{R}^{M' \times N'},
    \label{eq:normalized_correlation}
\end{equation}
where $\mathbf{C}$ is the normalized correlation matrix representing the similarity between two point clouds. Based on the correlation matrix, the correspondences between $\bXp$ and $\bYp$ can be directly calculated by using the largest correlation for each point.

The SE(3) transformation between the source and target point clouds can be then estimated using the superpoints correspondences with a weighted variant of the Kabsch-Umeyama algorithm~\cite{kabsch1976solution,88573}:
\begin{equation}
    \mathbf{\hat{R}, \hat{t}} = \argmin_{\mathbf{R}, \mathbf{t}} 
      \sum_i^{\min(N', M')}{w_i \norm {\mathbf{R} \mathbf{\hat{x}}_i + \mathbf{t} - \mathbf{\hat{y}}_i}}^2,
\label{eq:rigidTransform}
\end{equation}
where $w_i= \bC(\hat{\mathbf{x}}_i, \hat{\mathbf{y}}_i)$ is the correspondence weight, $\mathbf{\hat{x}}_i$, $\mathbf{\hat{y}}_i$ are the $i$-th pair of matched superpoints.
We use $\min(N', M')$ operation because corresponding points can only be found for the point cloud with the minimum number of superpoints.

Although we use the highly discriminative feature representations enhanced by the attention module, the correspondences are inevitably noisy. 
How to filter out the outliers (\ie, incorrect correspondences)? 
We show that the normalized correlation matrix $\bC$ obtained from superpoints matching in Eq. (\ref{eq:normalized_correlation}) contains \emph{rich information}, allowing us to effectively reject outliers for robust transformation matrix estimation.
Specifically, if $\hbx_i$ is similar to multiple superpoints, \eg, because of the repetitive patterns, its matching to $\hby_i$ tends to be unreliable.
Therefore, the normalized correlation score between them $w_i=\bC(\hat{\mathbf{x}}_i, \hat{\mathbf{y}}_i)$ will be low since $\bC$ is normalized w.r.t. all other superpoints in the target point cloud.
Unlike~\cite{xu2022gmflow}, which completely discards the correlation weights in $\bC$, we use $w_i$ for two purposes: discarding the unreliable corresondences that have low $w_i$ values and further weighting the rest inliers when fitting the transformation matrix in Eq. (\ref{eq:normalized_correlation}). 
It is important to note here that $w_i$ is not learned and is just the correlation score between the feature representations of $\mathbf{\hat{x}}_i$, $\mathbf{\hat{y}}_i$, the $i$-th pair of matched superpoints.
We show in the experiments that such a matching strategy works more effectively for superpoints matching than other approaches~\cite{tyszkiewicz2020disk, xu2022gmflow, sun2021loftr}.

In comparison to ~\cite{qin2022geometric, yu2021cofinet, 9859814}, our approach is \emph{simple yet effective}, which eliminates the coarse-to-fine strategy and more importantly, the inherently slow RANSAC-like postprocessing. %
Although the feature enhancement module is also used in~\cite{yew2022regtr}, our approach is fundamentally different. ~\cite{yew2022regtr} predicts corresponding points for both source and target point clouds whereas our approach \emph{removes the ambiguity of predicting non-existing points} and directly matches superpoints.
The limitation of \cite{yew2022regtr} can also be seen in outdoor LiDAR experiments like KITTI.
As shown in Table \ref{kitti-table}, it performs significantly worse due to predicting correspondences that do not align well with true corresponding points.

\subsection{Loss Functions}

We train our approach using the following three loss functions, where the transformation loss is the main loss term and the other two are auxiliary ones.

\mypar{Transformation Loss.}
We apply the L1 loss on the predicted transformed locations of all keypoints  with the predicted and ground truth transformation matrix. 
This is different from \cite{yew2022regtr}.
Since we do not predict corresponding points for each point cloud, we do not need to compute two transformation matrices and thus we do not have to force the network to learn that both matrices are inverses of each other.
\begin{align}
    \mathcal{L}_{T} = \frac{1}{M'} {\sum_i^{M'}} \left\lvert \hat{\mathbf{R}} \mathbf{x_i'} + \hat{\mathbf{t}} - \left(\mathbf{R}_{gt} \mathbf{x_i'} + \mathbf{t}_{gt}\right)\right\rvert_1.
\end{align}

\mypar{Overlap Loss.} Inspired by~\cite{yew2022regtr}, we estimate the overlap values $\mathbf{\hat{O}_{\mathbf{X'}}}$ and $\mathbf{\hat{O}_{\mathbf{Y'}}}$ using a separate MLP layer based on the enhanced feature $\bbF_{\bXp}$ and $\bbF_{\bYp}$, respectively.
The overlap estimation is formulated as a binary classification problem, so we use the binary cross-entropy loss:
\begin{align}
    \mathcal{L}^X_{o} = -\frac{1}{M'}\sum_{i}^{M'}{
        o_{x,i}^* \cdot \log \hat{o}_{x,i} + (1 - o_{x,i}^*) \cdot \log \left(1 - \hat{o}_{x,i}\right)}, 
\end{align}
where $\hat{o}_{x,i}$ is the estimated overlap probability and $o_{x,i}^*$ is the ground truth probability.
We compute the overlap loss $\mathcal{L}^Y_{o}$ for the target point cloud similarly.

\newcommand{\bvf}{\bar{\mathbf{f}}}
\newcommand{\vW}{\mathbf{W}}
\mypar{Feature Loss.}
Following~\cite{yew2022regtr}, to ensure that the enhanced features of both point clouds are in the same feature space, we apply an InfoNCE~\cite{infonce} loss on the enhanced features $\bbF_{\bXp}$ and $\bbF_{\bYp}$. 
Given a set of superpoints correspondences $\{(\hbx_i, \hby_i)\}_{i=1}^K$ and their associated feature representations $\{(\bvf_{\hbx_i}, \bvf_{\hby_i})$, the feature loss is defined as
\begin{equation}\label{eq:infonce}
    \mathcal{L}_{f} = -\frac{1}{K}
    \log
    \frac{\bvf_{\hbx_i}^T \vW \bvf_{\hby_i}}
    {\bvf_{\hbx_i}^T \vW \bvf_{\hby_i} + \sum_{j\neq i} \bvf_{\hbx_i}^T \vW \bvf_{\hby_j}}. 
\end{equation}
The linear transformation $\mathbf{W}$ is enforced to be symmetrical by parameterizing it as the sum of an upper triangular matrix $\mathbf{U}$ and its transpose, \ie $\mathbf{W} = \mathbf{U} + \mathbf{U}^T$.

The final loss is a weighted sum of all the losses with 
\begin{equation}
    \mathcal{L} = \mathcal{L}_{T} + \alpha \mathcal{L}_{f} + \beta (\mathcal{L}_{o}^X + \mathcal{L}_{o}^Y),
\end{equation}
where we set the loss weights $\alpha = 0.1$ and $\beta = 1$ empirically. 

\section{Experiments}
We evaluate our approach on three datasets with overlap ranging from 10\% to 75\%. The first dataset is on synthetic ModelNet dataset with two benchmarks settings following \cite{huang2021predator, yew2022regtr}.
The second one is 3DMatch \cite{zeng20173dmatch} with two benchmarks following \cite{huang2021predator, yew2022regtr, qin2022geometric, yu2021cofinet}.
The last one is on the challenging large-scale outdoor KITTI dataset \cite{Geiger2012CVPR}

\subsection{Implementation details}
Our approach is implemented using the PyTorch framework \cite{paszke2019pytorch} on a system with an Intel i9-1300K CPU and a single RTX 3090 GPU. The network training is performed with the AdamW optimizer \cite{loshchilov2018decoupled}, using a learning rate of 0.0001 and a weight decay of 0.0001. We tain the network for 400 epochs with batch size of 4 on ModelNet, 50 epochs with batch size of 4 on 3DMatch, and 80 epochs with batch size of 1 on KITTI.

\subsection{ModelNet and ModelLoNet Benchmarks}
The ModelNet40 \cite{wu20153d} dataset comprises of synthetic CAD models. Following the data setting in \cite{huang2021predator,yew2022regtr}, the point clouds are randomly sampled from mesh faces of the CAD models, cropped and subsampled.

Our network is trained exclusively on ModelNet, and evaluated for generalization on ModelLoNet. For benchmarking the performance of our model we use the Relative Rotation Error (RRE) and Relative translation Error (RTE) and Chamfer Distance (CD) as the primary metrics, following \cite{yew2022regtr}.

\begin{table}
\setlength\tabcolsep{2 pt}
  \caption{Registration results on ModelNet and ModelLoNet.}
  \label{modelnet-table}
  \centering
  \begin{tabular}{l|ccc|ccc}
    \toprule
    Methods & \multicolumn{3}{c|}{ModelNet} & \multicolumn{3}{c}{ModelLoNet}\\
    \cmidrule{2-7}
                         & RRE$\downarrow$     & RTE$\downarrow$     & CD$\downarrow$    & RRE$\downarrow$     & RTE$\downarrow$    & CD$\downarrow$  \\
    \midrule
    PNLK~\cite{aoki2019pointnetlk}          & 29.725      & 0.297   & 0.0235  & 48.567    & 0.507  & 0.0367 \\
    OMNet~\cite{xu2021omnet}               & 2.947       & 0.032   & 0.0015  & 6.517     & 0.129  & 0.0074 \\
    DCPv2~\cite{loshchilov2018decoupled}              & 11.975      & 0.171   & 0.0117  & 16.501    & 0.300  & 0.0268 \\
    RPMNet~\cite{yew2020rpm}             & 1.712       & 0.018   & 0.00085 & 7.342     & 0.124  & 0.0050 \\
    Predator~\cite{huang2021predator}           & 1.739       & 0.019   & 0.00089 & 5.235     & 0.132  & 0.0083 \\
    RegTR~\cite{yew2022regtr}               & \secondbest{1.473}       & \secondbest{0.014}   & \secondbest{0.00078} & \secondbest{3.930}     & \best{0.087}  & \best{0.0037}\\
    GeoT~\cite{qin2022geometric}      & 1.568       & 0.018   &   -      & \best{3.809}     & 0.102  &    -   \\
    \midrule
    Ours                & \best{1.247}      & \best{0.011}   & \best{0.00074} & \best{3.809}     & \secondbest{0.088}   & \secondbest{0.0040} \\
    \bottomrule
  \end{tabular}
\end{table}

The results are shown in Table \ref{modelnet-table}. We compare against correspondence-based approaches \cite{huang2021predator, yew2022regtr}, coarse-to-fine registration approaches \cite{qin2022geometric}, and end-to-end methods \cite{aoki2019pointnetlk, yew2020rpm, xu2021omnet}. Our approach performs well on both benchmarks improving results on ModelNet benchmark by 15\% in RRE and 21\% in RTE. The low chamfer error suggests that predicted correspondences have very high accuracy. Our approach is also able to outperform methods using post-processing steps like RANSAC \cite{huang2021predator} by a significant margin.

\subsection{3DMatch and 3DLoMatch Benchmarks}

3DMatch \cite{zeng20173dmatch} is a collection of 62 scenes, from which we use 46 for training, 8 for testing, and 8 for validation following \cite{yew2022regtr, huang2021predator, qin2022geometric}. We use the preprocessed data from \cite{huang2021predator} which contains point clouds downsampled using a voxel-grid subsampling method. The 3DMatch benchmark contains point clouds pairs with >30\% overlap while the 3DLoMatch benchmark contains scan pairs with only 10\%-30\% overlap. Following \cite{yew2022regtr}, we perform training data augmentation by applying small rigid perturbations, jittering, and shuffling of points. 
Following the literature \cite{huang2021predator, qin2022geometric, zeng20173dmatch}, We report the results of 3DMatch dataset on 5 metrics including RRE, RTE, Registration Recall (RR), Feature Matching Recall (FMR), and Inlier Ratio (IR). 

\begin{table}[t]
\setlength\tabcolsep{2 pt} 
  \caption{Registration results on  3DMatch and 3DLoMatch.}
  \label{3dmatch-table}
  \centering
  \begin{tabular}{l|ccccc|ccccc}
    \toprule
    \multirow{2}{*}{Methods} & \multicolumn{5}{c|}{3DMatch} & \multicolumn{5}{c}{3DLoMatch} \\
    \cmidrule{2-11}
                                    & RRE$\downarrow$     & RTE$\downarrow$     & RR$\uparrow$  & FMR$\uparrow$ & IR$\uparrow$    & RRE$\downarrow$     & RTE$\downarrow$    & RR$\uparrow$   & FMR$\uparrow$ & IR$\uparrow$ \\
    \midrule
    FCGF                            & 1.949     & 0.066     & 85.1    & 97.4  & 56.8 & 3.147     & 0.100     & 40.1 & 76.6    & 21.4  \\
    D3Feat                          & 2.161     & 0.067     & 81.6    & 95.6  & 39.0 & 3.361     & 0.103     & 37.2 & 67.3    & 13.2  \\
    Predator                        & 2.029     & 0.064     & 89.0    & 96.6  & 58.8 & 3.048     & 0.093     & 59.8 & 78.6    & 26.7  \\
    DGR                             & 2.103     & 0.067     & 85.3    &  -    & - & 3.954     & 0.113     & 48.7 &   -  & -  \\
    RegTR                           & \secondbest{1.567}     & \secondbest{0.049}     & \secondbest{92.0}    &  -   & - & 2.827     & \secondbest{0.077}&  64.8 &- & - \\  
    YOHO                            &  -   &  -    & 90.8    &  \best{98.2}    & 64.4 & -     & -     & 65.2 &   79.4  & 25.9 \\
    CofiNet                          &   -   &  -   & 89.3    &  \secondbest{98.1}   & 49.8 & -    & -     & \secondbest{67.5} &   \secondbest{83.1}  & 24.4  \\  
    GeoT          & 1.625     & 0.053     & 91.5      & 97.7   & \secondbest{70.3}   & \best{2.547}  & \best{0.074}  & \best{74.0}  & \best{88.1} & \secondbest{43.3}\\
    
    \midrule
    \textbf{Ours}             & \best{1.436} & \best{0.045} & \best{93.7}  & 96.5    & \best{89.8}    & \secondbest{2.553}     & \best{0.074}     & 65.0  & 76.5 & \best{57.5} \\
    \bottomrule
  \end{tabular}
\end{table}

We compare our approach against several learned correspondence-based algorithms \cite{gojcic2019perfect, choy2019fully, bai2020d3feat, huang2021predator} and coarse-to-fine approaches \cite{yu2021cofinet, qin2022geometric}. The quantitative results are shown in Table \ref{3dmatch-table}. For the 3DMatch benchmark, our approach outperforms all the previous methods in all but the FMR metric.
This implies that in cases of significant overlap (>30\%), the superpoint correspondences are very distinctive and accurate. For the 3DLoMatch benchmark, in comparison to all the approaches that do not use any post-processing, \eg Predator\cite{huang2021predator} and CoFiNet\cite{yu2021cofinet}, our method performs significantly better. GeoT~\cite{qin2022geometric} uses Local to Global Refinement (LGR) as the refinement step to get best results. In our case,  outliers are filtered by the correlation weights, thus providing accurate correspondences. The validity of the correspondences obtained by our approach can be verified by comparing the mean Inlier Ratio (IR) in Table~\ref{3dmatch-table}. We get the highest mean IR, almost \textbf{19\%} better than the second-best approach on 3DMatch and \textbf{14\%} higher in 3DLoMatch. 
Our approach does not perform the best in FMR because we prioritize outputting correct correspondences over a large number of, yet potentially noisy matchings. 
In contrast, CofiNet~\cite{yu2021cofinet}, has a very high FMR but low IR due to its propensity for producing many inaccurate matches. 

\begin{figure}[t]
\centering
    \begin{tabular}{cc}
       \includegraphics[width=0.45\linewidth]{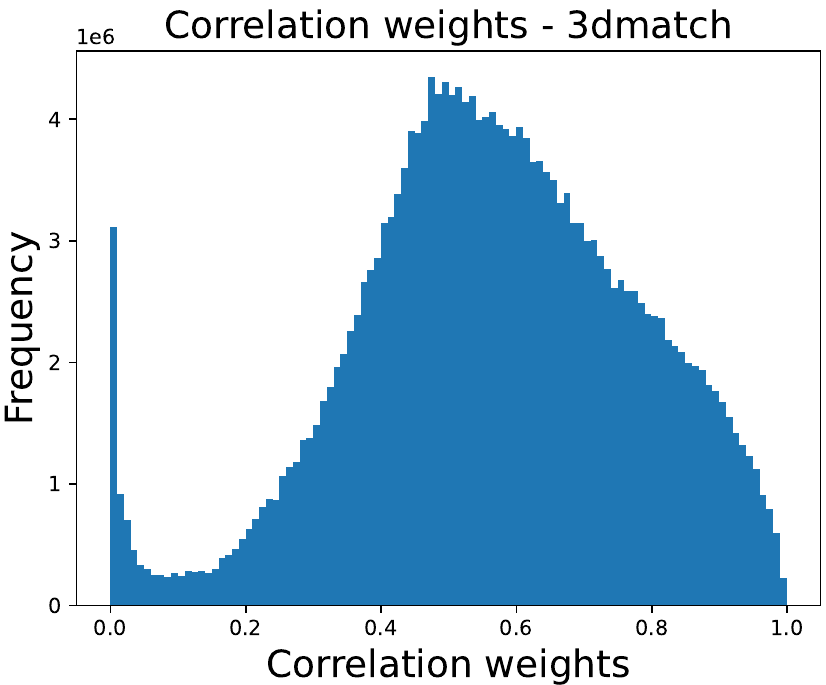} &
       \includegraphics[width=0.45\linewidth]{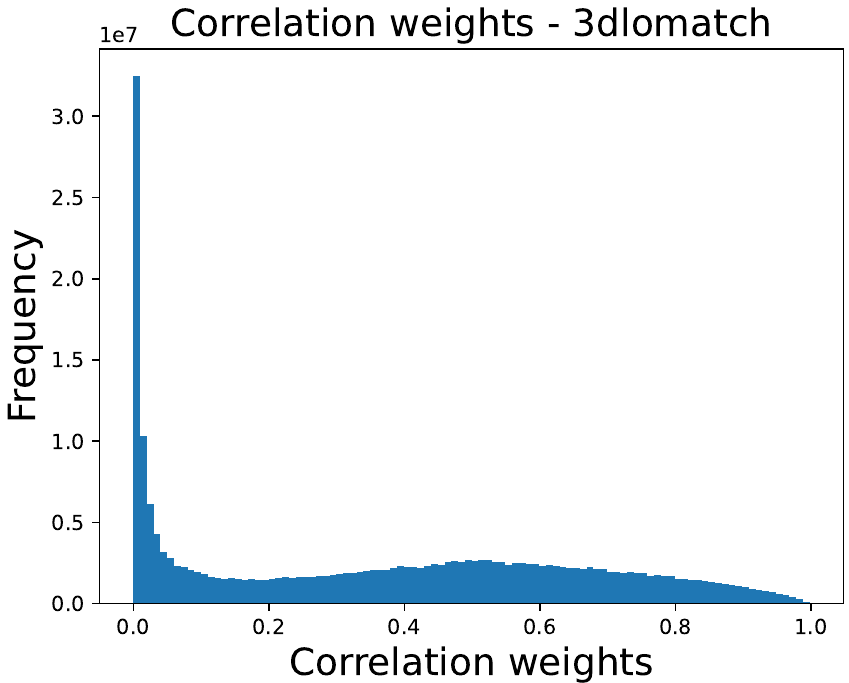}        
    \end{tabular}
   
   \caption{
       \textbf{Correlation weight distribution on 3DMatch.} Here, many points exhibit moderate correlations, suggesting that a considerable portion of sample points are reasonably good matches. In contrast, 3DLoMatch has fewer points that exhibit strong correlations due to the low data overlap.
       }
   \label{fig:wegiht_distribution}
\end{figure}

One of the problems associated with superpoint matching in point clouds is the resolution issue, where we might not have one-to-one correspondences due to subsampling. 
In our case, this is automatically handled by the correlation weights. The correlation weights are lower for weakly matching superpoints and thus even if there is no one-to-one correspondence, weak correspondences will be used according to their weights to compute the correct SE(3) transformation.
Figure \ref{fig:wegiht_distribution} shows the correlation weight distribution on the 3DMatch and 3DLoMatch benchmark. In 3DMatch, we have many high correlation weights,
suggesting that many sample points are good correspondences. On 3DLoMatch, we have much fewer good correlations suggesting that many of the sampled superpoints are not good matches, which is expected because of very low overlap.

\begin{figure}
    \centering
       \includegraphics[width=0.9\linewidth]{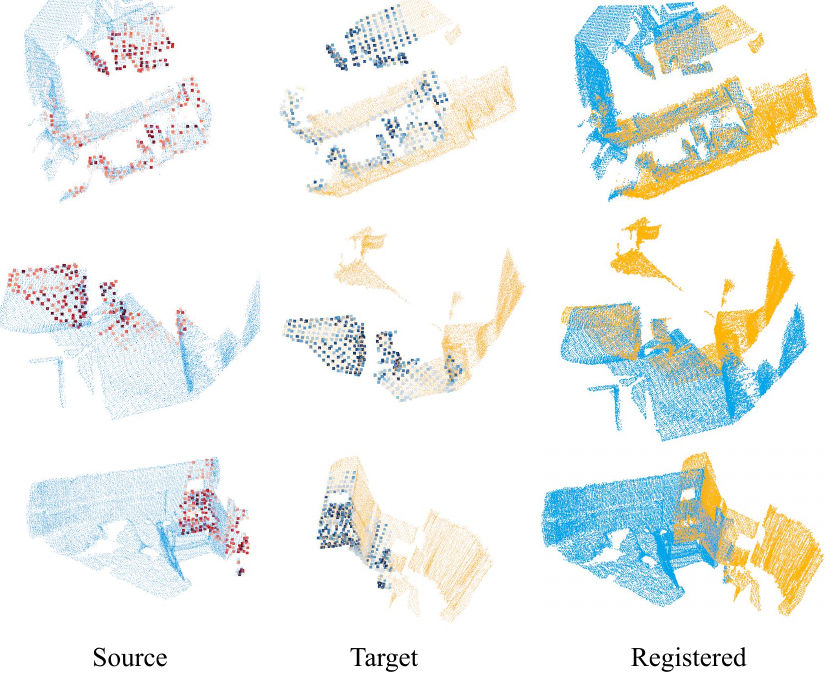}
       \caption{
           \textbf{Qualitative Results on 3DMatch.} The source cloud is shown in blue with the red color representing the matched superpoints. Similarily target cloud is shown in yellow color with matching superpoints in blue. (Best viewed in color.)
           }
       \label{fig:reg_res}
\end{figure}
\subsection{KITTI Benchmark}
KITTI odometry consists of 11 sequences of outdoor driving scenarios scanned using the HDL64 LiDAR sensor. Following ~\cite{huang2021predator}, we use sequence 0-5 for training, 6-7 for validation, and 8-10 for testing. As in ~\cite{huang2021predator, qin2022geometric} we only use the point cloud pairs that are at least 10m apart and their ground-truth poses are refined using ICP for all evaluations. 
Table \ref{kitti-table} shows the results obtained on KITTI. 
We can see that the baseline \cite{yew2022regtr} performs significantly worse in RTE. Since the correspondences are predicted, they do not align one-on-one with actual correspondences, thus significantly reducing registration accuracy.
We achieve state of the art results in RRE and RR while performing comparably in RTE. This showcases the effectiveness of our approach in large scale outdoor point clouds. Figure \ref{fig:teaser} shows an example of matched superpoints on KITTI dataset, our approach is able to find very distinctive correspondences to register the point cloud pair.

\begin{table}
  \caption{Registration results on KITTI.}
  \label{kitti-table}
  \centering
  \begin{tabular}{l|ccc}
    \toprule
    Methods                         & RTE$\downarrow$     & RRE$\downarrow$     & RR$\uparrow$  \\
    \midrule
    3DFeat~\cite{yew20183dfeat}                         &  25.9 & \secondbest{0.25} & 96.0 \\
    FCGF~\cite{rusu2009fast}                            &  9.5 & 0.30 & 96.6 \\
    D3Feat~\cite{bai2020d3feat}                          &  \secondbest{7.2} & 0.30 & \best{99.8} \\
    Predator~\cite{huang2021predator}                        &  \best{6.8} & 0.27 & \best{99.8} \\
    CofiNet~\cite{yu2021cofinet}                         &  8.2 & 0.41 & \best{99.8} \\
    RegTR~\cite{yew2022regtr}                           &  49.9 & 0.61 & \secondbest{99.1} \\
    GeoTransformer~\cite{qin2022geometric}                  &  \best{6.8} & \best{0.24} & \best{99.8} \\
    Unified BEV~\cite{li2023unified}                     &  7.5 & 0.26 & \best{99.8} \\
    \midrule
    \textbf{Ours}                   &  \secondbest{7.2} & \best{0.24} & \best{99.8}\\
    \bottomrule
  \end{tabular}
\end{table}

\subsection{Ablation Studies}

\begin{table}
\setlength\tabcolsep{2 pt}
\centering
\caption{Comparisons of different outlier filtering methods.}
\label{tab:outlier_filtering_methods}
\begin{tabular}{c|ccc|ccc|l}
    \toprule
    Outlier Filtering   & \multicolumn{3}{c|}{3DMatch} & \multicolumn{3}{c|}{3DLoMatch}\\
    \cmidrule{2-7}
       Method     & RRE       & RTE       & RR        & RRE       & RTE       & RR       & Time   \\
    \midrule 
      No filtering                      & \secondbest{1.462}     & \secondbest{0.046}     & \secondbest{93.4}      & \secondbest{2.652}     & \best{0.074}     & \secondbest{64.6}       & \best{0.073}\\
      Ours + RANSAC                 & 2.788     & 0.090     & 82.3      & 4.653     & 0.115     & 32.0          & \secondbest{0.141}\\
      Ours (top 15\%) + RANSAC      & 1.701     & 0.051     & 93.0      & 2.696     & \secondbest{0.078}     & 64.5          & 0.153\\
      Correlation scores (top 15\%)     & \best{1.436}     & \best{0.045}     & \best{93.7}      & \best{2.553}     & \best{0.074}     & \best{65.0} & \best{0.073}\\
    \bottomrule
\end{tabular}
\end{table}

\begin{table}[b]
\caption{Effectiveness of different loss terms.}
\label{tab:loss_terms}
\begin{tabular}{cc|lll|lll}
    \toprule
    Overlap & Feature & \multicolumn{3}{c|}{3DMatch} & \multicolumn{3}{c}{3DLoMatch}\\
    \cmidrule{3-8}
     Loss & Loss      & RRE       & RTE       & RR        & RRE       & RTE       & RR   \\
    \midrule 
      \xmark & \xmark  & 2.521     & 0.076     & 76.5      & 5.272     & 0.132     & 31.2\\
      \cmark & \xmark  & 2.123     & 0.062        &79.6       & 4.020   & 0.105 &37.2\\
      \xmark & \cmark  & 1.651      & 0.049     & 90.3      & 2.876     & 0.082     & 58.0 \\
      \cmark & \cmark  & \best{1.436} & \best{0.045}  & \best{93.7}   & \best{2.553}  & \best{0.074}  & \best{65.0}\\
    \bottomrule
\end{tabular}
\end{table}

\textbf{Using correlation scores for filtering outliers.} 
We study the effectiveness of outlier filtering in Table~\ref{tab:outlier_filtering_methods}.
In the `No filtering' approach, we use all the superpoints with their scores \emph{without any filtering}. A little surprisingly, it achieves very competitive results, likely because we use the correlation weights of the matchings in the transformation matrix estimation in Eq.(\ref{eq:rigidTransform}).
If the outlier filtering is used (the last row), the best accuracy can be obtained.

We also experiment with RANSAC on the set of superpoints correspondences with and without outlier filtering, respectively.
Without any filtering (`Ours + RANSAC'), the large number of outliers impose significant challenges for RANSAC to find the optimal solution.
Even with filtering (`Ours (top 15\%) + RANSAC'), we can see that RANSAC gives slightly worse results, partially because it does not always converge to the optimal solution.
Furthermore, RANSAC requires twice as much time for the estimation of the transformation matrix compared to our method.

\textbf{Effectiveness of the loss terms.} %
We also analyze the effectiveness of each loss function. Table \ref{tab:loss_terms} shows the results with different loss function configurations. We see that by just using the feature loss along with the transformation loss, the model is able to achieve good performance.
Using the overlap loss further helps the model prune remaining outliers.

\textbf{Matching strategy.}
The choice of a matching strategy is crucial for precise pose estimation. In prior approaches, such as \cite{qin2022geometric, yew2020rpm, yu2021cofinet}, the Sinkhorn algorithm has commonly been employed to compute optimal transport and derive matching scores for correspondences. In our research, we opt for Global Softmax, which, as evidenced by our experiments, outperforms the Sinkhorn approach (refer to Table \ref{sinkhorn comparison}). 
Note that while Sinkhorn algorithm requires careful parameter tuning, our approach is parameter-free and works out of the box.
While Global Softmax has been previously used as a matching strategy in \cite{tyszkiewicz2020disk, yew2020rpm}, prior works only utilized it to identify matching points with probabilities exceeding a certain threshold, resulting in suboptimal outcomes. Our innovation lies in utilizing these probabilities as correspondence weights, facilitating gradient flow through the Weighted Kabsch-Umeyama Solver which leads to faster optimization and better results. We also experimented with Dual Softmax matching used in LofTR~\cite{sun2021loftr}, but the difference in performance is less than 1\% with Global Softmax being slightly better 
and faster than Dual-Softmax.

\label{sec:match_strat}
\begin{table}[t]
\setlength\tabcolsep{2 pt} 
  \caption{Comparison of Matching Strategy}
  \label{sinkhorn comparison}
  \centering
  \begin{tabular}{l|ccc|ccc|l}
    \toprule
    \multirow{2}{*}{Methods} & \multicolumn{3}{c|}{3DMatch} & \multicolumn{3}{c|}{3DLoMatch} & Time \\
    \cmidrule{2-7} & RRE$\downarrow$     & RTE$\downarrow$     & RR$\uparrow$    & RRE$\downarrow$     & RTE$\downarrow$    & RR$\uparrow$ & \\
    \midrule
    Sinkhorn            &  1.509 & 0.046 & 89.2 & 2.622 & \best{0.074} & 59.6 & 0.094\\
    Dual Softmax        & 1.441 & \best{0.045} & 93.6 & 2.589 & \best{0.074} & \best{65.1} & 0.080 \\
    MLP \cite{yew2022regtr}  & 1.567 & 0.049 & 92.0 & 2.827 & 0.077 & 64.8 & 0.078 \\
    Global Softmax (Ours)      &  \best{1.436} & \best{0.045} & \best{93.7} & \best{2.553} & \best{0.074} & 65.0 & \best{0.073}\\
    \bottomrule
  \end{tabular}
\end{table}

\textbf{Data Augmentation Schemes}
Previous works have tried different data augmentation schemes. ~\cite{yew2022regtr} uses a weak data augmentation, perturbing poses by a small amount during training. While ~\cite{qin2022geometric, yew2020rpm, huang2021predator} use large data augmentation, perturbing poses by a full range of motion. Weak data augmentation can slightly improve performance but deteriorates network generalization. 
Table \ref{tab: data_augmentation_comparison} shows that our approach achieves slightly better performance with weak data augmentation. 
But with strong data augmentation, the generalization on 3DLoMatch benchmarks improves RR by 1.8\%. 

\begin{table}[t]
\setlength\tabcolsep{2 pt} 
  \caption{Comparison of Data Augmentation Schemes}
  \label{tab: data_augmentation_comparison}
  \centering
  \begin{tabular}{l|ccc|ccc}
    \toprule
    \multirow{2}{*}{Methods} & \multicolumn{3}{c|}{3DMatch} & \multicolumn{3}{c}{3DLoMatch} \\
    \cmidrule{2-7} & RRE$\downarrow$     & RTE$\downarrow$     & RR$\uparrow$    & RRE$\downarrow$     & RTE$\downarrow$    & RR$\uparrow$  \\
    \midrule
     Ours (Weak Augmentation) &  \best{1.436} & \best{0.045} & \best{93.7} & \best{2.553} & \best{0.074} & 65.0\\
     Ours (Large Augmentation)    & 1.472 & \best{0.045} & 93.4 & 2.611 & \best{0.074} & \best{66.8} \\
    \bottomrule
  \end{tabular}
\end{table}

\section{Conclusion}
In this paper, we presented a strong baseline approach for point cloud registration. By using Global Softmax to directly match superpoint features, we remove the ambiguity of predicting non-existent corresponding points while using the softmax probabilities as correspondence weights allows us to filter outliers without any post-processing refinement. 
Experimental results on standard benchmarks show that our model achieves comparable or even better accuracy than state-of-the-art methods.
We do not advocate our approach to be \emph{the} solution for point cloud registration.
Rather, we'd like to emphasize the role of matching strategy for point cloud registration, showcasing the feasibility of achieving high accuracy without cumbersome ad hoc postprocessing.

{
\bibliographystyle{plain}
\bibliography{main.bib}

\begin{thebibliography}{10}

\bibitem{ao2021spinnet}
Sheng Ao, Qingyong Hu, Bo~Yang, Andrew Markham, and Yulan Guo.
\newblock Spinnet: Learning a general surface descriptor for 3d point cloud registration.
\newblock In {\em CVPR}, 2021.

\bibitem{aoki2019pointnetlk}
Yasuhiro Aoki, Hunter Goforth, Rangaprasad~Arun Srivatsan, and Simon Lucey.
\newblock Pointnetlk: Robust \& efficient point cloud registration using pointnet.
\newblock In {\em CVPR}, 2019.

\bibitem{bai2021pointdsc}
Xuyang Bai, Zixin Luo, Lei Zhou, Hongkai Chen, Lei Li, Zeyu Hu, Hongbo Fu, and Chiew-Lan Tai.
\newblock Pointdsc: Robust point cloud registration using deep spatial consistency.
\newblock In {\em CVPR}, 2021.

\bibitem{bai2020d3feat}
Xuyang Bai, Zixin Luo, Lei Zhou, Hongbo Fu, Long Quan, and Chiew-Lan Tai.
\newblock D3feat: Joint learning of dense detection and description of 3d local features.
\newblock In {\em CVPR}, 2020.

\bibitem{besl1992method}
Paul~J Besl and Neil~D McKay.
\newblock Method for registration of 3-d shapes.
\newblock In {\em Sensor fusion IV: control paradigms and data structures}, 1992.

\bibitem{bouaziz2013sparse}
Sofien Bouaziz, Andrea Tagliasacchi, and Mark Pauly.
\newblock Sparse iterative closest point.
\newblock In {\em Computer graphics forum}, 2013.

\bibitem{brachmann2017dsac}
Eric Brachmann, Alexander Krull, Sebastian Nowozin, Jamie Shotton, Frank Michel, Stefan Gumhold, and Carsten Rother.
\newblock Dsac-differentiable ransac for camera localization.
\newblock In {\em CVPR}, 2017.

\bibitem{cao2021pcam}
Anh-Quan Cao, Gilles Puy, Alexandre Boulch, and Renaud Marlet.
\newblock Pcam: Product of cross-attention matrices for rigid registration of point clouds.
\newblock In {\em ICCV}, 2021.

\bibitem{chetverikov2002trimmed}
Dmitry Chetverikov, Dmitry Svirko, Dmitry Stepanov, and Pavel Krsek.
\newblock The trimmed iterative closest point algorithm.
\newblock In {\em ICPR}, 2002.

\bibitem{choy2020deep}
Christopher Choy, Wei Dong, and Vladlen Koltun.
\newblock Deep global registration.
\newblock In {\em CVPR}, 2020.

\bibitem{choy2019fully}
Christopher Choy, Jaesik Park, and Vladlen Koltun.
\newblock Fully convolutional geometric features.
\newblock In {\em CVPR}, 2019.

\bibitem{deng2018ppf}
Haowen Deng, Tolga Birdal, and Slobodan Ilic.
\newblock Ppf-foldnet: Unsupervised learning of rotation invariant 3d local descriptors.
\newblock In {\em ECCV}, 2018.

\bibitem{deng2018ppfnet}
Haowen Deng, Tolga Birdal, and Slobodan Ilic.
\newblock Ppfnet: Global context aware local features for robust 3d point matching.
\newblock In {\em CVPR}, 2018.

\bibitem{Deschaud2018imlsslam}
Jean{-}Emmanuel Deschaud.
\newblock {IMLS-SLAM:} scan-to-model matching based on 3d data.
\newblock In {\em {ICRA}}, 2018.

\bibitem{fischler1981random}
Martin~A Fischler and Robert~C Bolles.
\newblock Random sample consensus: a paradigm for model fitting with applications to image analysis and automated cartography.
\newblock {\em Communications of the ACM}, 1981.

\bibitem{Geiger2012CVPR}
Andreas Geiger, Philip Lenz, and Raquel Urtasun.
\newblock Are we ready for autonomous driving? the kitti vision benchmark suite.
\newblock In {\em Conference on Computer Vision and Pattern Recognition (CVPR)}, 2012.

\bibitem{gojcic2020learning}
Zan Gojcic, Caifa Zhou, Jan~D Wegner, Leonidas~J Guibas, and Tolga Birdal.
\newblock Learning multiview 3d point cloud registration.
\newblock In {\em CVPR}, 2020.

\bibitem{gojcic2019perfect}
Zan Gojcic, Caifa Zhou, Jan~D Wegner, and Andreas Wieser.
\newblock The perfect match: 3d point cloud matching with smoothed densities.
\newblock In {\em CVPR}, 2019.

\bibitem{Gross193DV}
Johannes Gro{\ss}, Aljo\u{s}a O\u{s}ep, and Bastian Leibe.
\newblock Alignnet-3d: Fast point cloud registration of partially observed objects.
\newblock In {\em 3DV}, 2019.

\bibitem{he2016deep}
Kaiming He, Xiangyu Zhang, Shaoqing Ren, and Jian Sun.
\newblock Deep residual learning for image recognition.
\newblock In {\em CVPR}, 2016.

\bibitem{huang2021predator}
Shengyu Huang, Zan Gojcic, Mikhail Usvyatsov, Andreas Wieser, and Konrad Schindler.
\newblock Predator: Registration of 3d point clouds with low overlap.
\newblock In {\em CVPR}, 2021.

\bibitem{Izadi2011kinect}
Shahram Izadi, Richard~A. Newcombe, David Kim, Otmar Hilliges, David Molyneaux, Steve Hodges, Pushmeet Kohli, Jamie Shotton, Andrew~J. Davison, and Andrew~W. Fitzgibbon.
\newblock Kinectfusion: real-time dynamic 3d surface reconstruction and interaction.
\newblock In {\em {SIGGRAPH}}, 2011.

\bibitem{kabsch1976solution}
Wolfgang Kabsch.
\newblock A solution for the best rotation to relate two sets of vectors.
\newblock {\em Acta Crystallographica Section A: Crystal Physics, Diffraction, Theoretical and General Crystallography}, 1976.

\bibitem{KhouryLCGF}
Marc Khoury, Qian-Yi Zhou, and Vladlen Koltun.
\newblock Learning compact geometric features.
\newblock In {\em ICCV}, 2017.

\bibitem{lee2021deep}
Junha Lee, Seungwook Kim, Minsu Cho, and Jaesik Park.
\newblock Deep hough voting for robust global registration.
\newblock In {\em ICCV}, 2021.

\bibitem{li2019usip}
Jiaxin Li and Gim~Hee Lee.
\newblock Usip: Unsupervised stable interest point detection from 3d point clouds.
\newblock {\em ICCV}, 2019.

\bibitem{li2023unified}
Lin Li, Wendong Ding, Yongkun Wen, Yufei Liang, Yong Liu, and Guowei Wan.
\newblock A unified bev model for joint learning of 3d local features and overlap estimation, 2023.

\bibitem{li2021deep}
Ying Li, Lingfei Ma, Zilong Zhong, Fei Liu, Michael~A. Chapman, Dongpu Cao, and Jonathan Li.
\newblock Deep learning for lidar point clouds in autonomous driving: A review.
\newblock {\em IEEE Transactions on Neural Networks and Learning Systems}, 2021.

\bibitem{SIFT}
Tony Lindeberg.
\newblock {\em Scale Invariant Feature Transform}, volume~7.
\newblock 05 2012.

\bibitem{liu2018efficient}
Yinlong Liu, Chen Wang, Zhijian Song, and Manning Wang.
\newblock Efficient global point cloud registration by matching rotation invariant features through translation search.
\newblock In {\em ECCV}, 2018.

\bibitem{loshchilov2018decoupled}
Ilya Loshchilov and Frank Hutter.
\newblock Decoupled weight decay regularization.
\newblock In {\em ICLR}, 2019.

\bibitem{Lu2019L3NetTL}
Weixin Lu, Yao Zhou, Guowei Wan, Shenhua Hou, and Shiyu Song.
\newblock L3-net: Towards learning based lidar localization for autonomous driving.
\newblock In {\em CVPR}, 2019.

\bibitem{9859814}
Guofeng Mei, Xiaoshui Huang, Jian Zhang, and Qiang Wu.
\newblock Overlap-guided coarse-to-fine correspondence prediction for point cloud registration.
\newblock In {\em ICME}, 2022.

\bibitem{infonce}
Aaron van~den Oord, Yazhe Li, and Oriol Vinyals.
\newblock Representation learning with contrastive predictive coding.
\newblock {\em arXiv}, 2018.

\bibitem{pais20203dregnet}
G~Dias Pais, Srikumar Ramalingam, Venu~Madhav Govindu, Jacinto~C Nascimento, Rama Chellappa, and Pedro Miraldo.
\newblock 3dregnet: A deep neural network for 3d point registration.
\newblock In {\em CVPR}, 2020.

\bibitem{paszke2019pytorch}
Adam Paszke, Sam Gross, Francisco Massa, Adam Lerer, James Bradbury, Gregory Chanan, Trevor Killeen, Zeming Lin, Natalia Gimelshein, Luca Antiga, et~al.
\newblock Pytorch: An imperative style, high-performance deep learning library.
\newblock {\em NeurIPS}, 2019.

\bibitem{Qi2017piontnet}
Charles~Ruizhongtai Qi, Hao Su, Kaichun Mo, and Leonidas~J. Guibas.
\newblock Pointnet: Deep learning on point sets for 3d classification and segmentation.
\newblock In {\em CVPR}, 2017.

\bibitem{Qi2017pointnetpp}
Charles~Ruizhongtai Qi, Li~Yi, Hao Su, and Leonidas~J. Guibas.
\newblock Pointnet++: Deep hierarchical feature learning on point sets in a metric space.
\newblock In {\em NeurIPS}, 2017.

\bibitem{qin2022geometric}
Zheng Qin, Hao Yu, Changjian Wang, Yulan Guo, Yuxing Peng, and Kai Xu.
\newblock Geometric transformer for fast and robust point cloud registration.
\newblock In {\em CVPR}, 2022.

\bibitem{rusinkiewicz2019symmetric}
Szymon Rusinkiewicz.
\newblock A symmetric objective function for icp.
\newblock {\em ACM Transactions on Graphics (TOG)}, 2019.

\bibitem{rusinkiewicz2001efficient}
Szymon Rusinkiewicz and Marc Levoy.
\newblock Efficient variants of the icp algorithm.
\newblock In {\em Proceedings third international conference on 3-D digital imaging and modeling}, 2001.

\bibitem{rusu2009fast}
Radu~Bogdan Rusu, Nico Blodow, and Michael Beetz.
\newblock Fast point feature histograms (fpfh) for 3d registration.
\newblock In {\em ICRA}, 2009.

\bibitem{segal2009generalized}
Aleksandr Segal, Dirk Haehnel, and Sebastian Thrun.
\newblock Generalized-icp.
\newblock In {\em Robotics: science and systems}, 2009.

\bibitem{legoloam2018}
Tixiao Shan and Brendan Englot.
\newblock Lego-loam: Lightweight and ground-optimized lidar odometry and mapping on variable terrain.
\newblock In {\em (IROS)}, 2018.

\bibitem{sinkhorn1967concerning}
Richard Sinkhorn and Paul Knopp.
\newblock Concerning nonnegative matrices and doubly stochastic matrices.
\newblock {\em Pacific Journal of Mathematics}, 1967.

\bibitem{sun2021loftr}
Jiaming Sun, Zehong Shen, Yuang Wang, Hujun Bao, and Xiaowei Zhou.
\newblock Loftr: Detector-free local feature matching with transformers.
\newblock In {\em CVPR}, 2021.

\bibitem{thomas2019kpconv}
Hugues Thomas, Charles~R Qi, Jean-Emmanuel Deschaud, Beatriz Marcotegui, Fran{\c{c}}ois Goulette, and Leonidas~J Guibas.
\newblock Kpconv: Flexible and deformable convolution for point clouds.
\newblock In {\em ICCV}, 2019.

\bibitem{tombari2010unique}
Federico Tombari, Samuele Salti, and Luigi Di~Stefano.
\newblock Unique signatures of histograms for local surface description.
\newblock In {\em ECCV}, 2010.

\bibitem{tyszkiewicz2020disk}
Micha{\l}~J Tyszkiewicz, Pascal Fua, and Eduard Trulls.
\newblock Disk: Learning local features with policy gradient.
\newblock {\em arXiv preprint arXiv:2006.13566}, 2020.

\bibitem{88573}
S.~Umeyama.
\newblock Least-squares estimation of transformation parameters between two point patterns.
\newblock {\em TPAMI}, 1991.

\bibitem{vaswani2017attention}
Ashish Vaswani, Noam Shazeer, Niki Parmar, Jakob Uszkoreit, Llion Jones, Aidan~N Gomez, {\L}ukasz Kaiser, and Illia Polosukhin.
\newblock Attention is all you need.
\newblock In {\em NeurIPS}, 2017.

\bibitem{wang2022you}
Haiping Wang, Yuan Liu, Zhen Dong, and Wenping Wang.
\newblock You only hypothesize once: Point cloud registration with rotation-equivariant descriptors.
\newblock In {\em ACM MM}, 2022.

\bibitem{wei2023generalized}
Tong Wei, Yash Patel, Alexander Shekhovtsov, Jiri Matas, and Daniel Barath.
\newblock Generalized differentiable ransac, 2023.

\bibitem{wu20153d}
Zhirong Wu, Shuran Song, Aditya Khosla, Fisher Yu, Linguang Zhang, Xiaoou Tang, and Jianxiong Xiao.
\newblock 3d shapenets: A deep representation for volumetric shapes.
\newblock In {\em CVPR}, 2015.

\bibitem{xu2021omnet}
Hao Xu, Shuaicheng Liu, Guangfu Wang, Guanghui Liu, and Bing Zeng.
\newblock Omnet: Learning overlapping mask for partial-to-partial point cloud registration.
\newblock In {\em ICCV}, 2021.

\bibitem{xu2022gmflow}
Haofei Xu, Jing Zhang, Jianfei Cai, Hamid Rezatofighi, and Dacheng Tao.
\newblock Gmflow: Learning optical flow via global matching.
\newblock In {\em CVPR}, 2022.

\bibitem{yew20183dfeat}
Zi~Jian Yew and Gim~Hee Lee.
\newblock 3dfeat-net: Weakly supervised local 3d features for point cloud registration.
\newblock In {\em ECCV}, 2018.

\bibitem{yew2020rpm}
Zi~Jian Yew and Gim~Hee Lee.
\newblock Rpm-net: Robust point matching using learned features.
\newblock In {\em CVPR}, 2020.

\bibitem{yew2022regtr}
Zi~Jian Yew and Gim~hee Lee.
\newblock Regtr: End-to-end point cloud correspondences with transformers.
\newblock In {\em CVPR}, 2022.

\bibitem{yi2018learning}
Kwang~Moo Yi, Eduard Trulls, Yuki Ono, Vincent Lepetit, Mathieu Salzmann, and Pascal Fua.
\newblock Learning to find good correspondences.
\newblock In {\em CVPR}, 2018.

\bibitem{yu2021cofinet}
Hao Yu, Fu~Li, Mahdi Saleh, Benjamin Busam, and Slobodan Ilic.
\newblock Cofinet: Reliable coarse-to-fine correspondences for robust pointcloud registration.
\newblock {\em NeurIPS}, 2021.

\bibitem{yuan2020deepgmr}
Wentao Yuan, Benjamin Eckart, Kihwan Kim, Varun Jampani, Dieter Fox, and Jan Kautz.
\newblock Deepgmr: Learning latent gaussian mixture models for registration.
\newblock In {\em ECCV}, 2020.

\bibitem{zeng20173dmatch}
Andy Zeng, Shuran Song, Matthias Nie{\ss}ner, Matthew Fisher, Jianxiong Xiao, and Thomas Funkhouser.
\newblock 3dmatch: Learning local geometric descriptors from rgb-d reconstructions.
\newblock In {\em CVPR}, 2017.

\end{thebibliography}
}

\end{document}